\documentclass[runningheads]{llncs}
\usepackage{comment}
\usepackage{xcolor}
\usepackage{graphicx}
\usepackage{multicol}
\usepackage{amssymb}
\usepackage{algpseudocode}
\usepackage{scalefnt}
\usepackage[utf8]{inputenc}  
\usepackage{booktabs}
\usepackage{helvet} 
\usepackage[pdftex]{hyperref}
\usepackage{enumerate} 
\usepackage{hyperref}
\usepackage{fancyhdr} 
\usepackage{scalefnt}

\begin{document}

\title{Studying Dishonest Intentions in Brazilian Portuguese Texts}
\titlerunning{Studying Dishonest Intentions in Brazilian Portuguese Texts}

\author{Francielle Alves Vargas \and
Thiago Alexandre Salgueiro Pardo } 
\authorrunning{Vargas, Cardoso and Pardo}

\author{Francielle Alves Vargas \inst{}\orcidID{0000-0003-3164-2682} \and
\\Thiago Alexandre Salgueiro Pardo \inst{}\orcidID{0000-0003-2111-1319}}
\institute{Interinstitutional Center for Computational Linguistics (NILC) \\
Institute of Mathematical and Computer Sciences, University of S\~ao Paulo\\
S\~ao Carlos, Brazil\\
\email{francielleavargas@usp.br}, \email{taspardo@icmc.usp.br}}

\maketitle             

\vspace{-4ex}
\begin{abstract}
Previous work in the social sciences, psychology and linguistics has show that liars have some control over the content of their stories, however their underlying state of mind may "leak out" through the way that they tell them. To the best of our knowledge, no previous systematic effort exists in order to describe and model deception language for Brazilian Portuguese. To fill this important gap, we carry out an initial empirical linguistic study on false statements in Brazilian news. We methodically analyze linguistic features using a deceptive news corpus, which includes both fake and true news. The results show that they present substantial lexical, syntactic and semantic variations, as well as punctuation and emotion distinctions.

\end{abstract}
\vspace{-4ex}

\section{Introduction}
According to the standard philosophical definition, lying is saying something that you believe to be false with the intent to deceive \cite{fallis2006}. For deception detection, the FBI trains its agents in a technique named statement analysis, which attempts to detect deception based on parts of speech (i.e., linguistics style) rather than the facts of the case or the story as a whole \cite{Adams1996}. This method is employed in interrogations, where the suspects are first asked to make a written statement. In \cite{NewmanetAll2003}, the authors report an example proposed by \cite{Adams1996} of a man accused of killing his wife. In this statement, the accused consistently refers to ``my wife and I'' rather than ``we'', suggesting distance between the couple. Thus, for \cite{NewmanetAll2003}, linguistic style checking may be useful in the hands of a trained expert who knows what to look for and how to use language to reveal inconsistencies. 

In this context, the deception spread through fake news and reviews is a relevant current problem. Due to their appealing nature, they spread rapidly \cite{Vosoughi2018}. Nevertheless, what makes fake content a hard problem to solve is the difficulty in identifying unreliable content. Fake news detection is defined as the prediction of the chances of a particular news article being intentionally deceptive \cite{Rubin:2015:DDN:2857070.2857153}, and fake reviews or opinion spam are inappropriate or fraudulent reviews \cite{ott-etal-2011-finding}.

 The psychologists and other social scientists are working hard to understand what drive people to believe in fake news. Unfortunately, there is not yet a consensus on this issue. As claimed by \cite{Pennycook&Rand2019}, much of the debate among researchers falls into two opposing camps. One group claims that our ability to reason is hijacked by our partisan convictions. The other group claims that the problem is that we often fail to exercise our critical faculties: that is, we are mentally lazy. 
 
 The rationalization camp, which has gained considerable prominence in recent years, is built around a set of theories contending that, when it comes to politically charged issues, people use their intellectual abilities to persuade themselves to believe in what they want to be true, rather than attempting to actually discover the truth. In the context of social media, some evidences suggest that the main factor explaining the acceptance of fake news could be cognitive laziness, especially, where news items are often skimmed or merely glanced at.
 
\cite{Finney2015} calls attention to a lack of non-laboratory studies. In \cite{Mann&Vrij2001}, the authors comment that their study, examining the deceptive and truthful statements of a convicted murderer, was, at the time, the only known study of its type in a ``high-stakes realistic setting''. Moreover, as believed by \cite{Meibauer2018}, we do not know much about the embedded lies in texts or discourses. With the notable exception of a paper published by \cite{Galasiski2000} and several studies proposed by \cite{Meibauer&Dynel} dealing with fictional discourse in the American television show, there is a lack of empirical research.

Therefore, in this paper, we present a pioneering empirical linguistic study for Brazilian Portuguese language on false statements in texts. We methodically analyze linguistic features from the Fake.Br corpus, which includes both fake and true news. The goal in the linguist approach is to investigate predictive deception clues found in texts. In particular, in this paper, we aim to provide linguistically motivated resources and computationally useful strategies for the development of automatic deception detection classifiers for the Portuguese language.

The remainder of this paper is organized as follows. In Section 2, we present the main related work. Section 3 describes an overview of our data. In Section 4, we show the entire empirical linguistic-based study. In Section 5, final remarks and future works are presented.

\section{Related Work}
\cite{DePauloetAl} defines deception as a deliberate attempt to mislead others. There are relatively few studies that have focused, specifically, on deceptive language recognition with speech or writing style, specially for Portuguese. Most of the available works have been used to aid in authorship attribution and plagiarism identification \cite{Cristani:2012:CSF:2393347.2396398}. Recent studies have been valuable for detecting deception, especially in the Fake News classification.

\cite{NewmanetAll2003} examined lying in written communication, finding that deceptive utterances used more total words but fewer personal pronouns. The linguistic-based features have been employed for fake news detection. In \cite{NewmanetAll2003}, the authors listed a set of linguistic behaviors that predict deception, as tones of words, kinds of preposition, conjunctions and pronouns. In addition, the deception linguistic style includes weak employment of singular and third person pronouns, negative polarity and frequent use of movement verbs. \cite{DePauloetAl} also presents a long study on clues to deception. For \cite{NahariEtAll2019}, the basic assumption is that liars differ from truth tellers in their verbal behaviour, making it possible to classify the news by inspecting their verbal accounts. Accordingly, they present insights, decisions, and conclusions resulting from deception research conference at legal and criminologist psychology society. In \cite{CoroyetAll2015}, the authors proposed a set of features using several linguistic analysis levels. They employed lexical, syntax, semantic and discourse linguistic features. In the lexical level, the authors explored bag of words (BWO) approach using bi-grams. In the syntax level, a probability context free grammar was implemented. For semantic analysis, the context information (such as profile content) has been incorporated. To model discourse features, the authors used the Rhetorical Structure Theory (RST) analytical framework.

Specifically for Brazilian Portuguese, \cite{SilvaEtAl2020} created the Fake.Br corpus and proposed classifiers for fake news detection. However, to the best of our knowledge, no previous systematic empirical linguistic study exists on dishonest intentions and language-based deception detection for the Brazilian Portuguese language.

\section{Data Overview}

To provide a linguistic analysis on false statements in texts, the first challenges concentrate on the data. The identification of reliable corpora for each language is a relevant task. Most of the research has developed computational linguistic resources for English. In general, few resources are available for Portuguese. As we commented before, for Brazilian Portuguese, we have Fake.Br \cite{SilvaEtAl2020}, which includes fake and true news in Brazilian Portuguese. An overview of this corpus is shown in Tables 1, 2 and 3.

\begin{table}[!h]
\caption{Corpus Overview: Fake.Br} 
\centering
\scalefont{0.85} 
\begin{tabular}{|p{30mm}|p{25mm}|p{8mm}|}
\hline
\textbf{Subjects}       & \textbf{Number of Texts}  & \textbf{\%} \\
\hline
Politics                & 4,180         & 58.0  \\
TV \& celebrities 	    & 1,544         & 21.4 \\
Society \& daily news 	& 1,276         & 17.7 \\
Science \& technology   & 112           & 1.5 \\
Economy                 & 44            & 0.7 \\
Religion                & 44            & 0.7 \\
\hline
\end{tabular}
\end{table}

 \begin{table}[!htb]
 \begin{minipage}{0.5\textwidth}
 \centering
 \scalefont{0.85} 
\caption{Number of tokens} 
\centering
\begin{tabular}{|p{15mm}|p{15mm}|p{10mm}|}
\hline
\textbf{News}    & \textbf{Tokens} & \textbf{\%}\\
\hline
Fake    & 796.364 & 50.80\\
True  	& 771.510 & 49.20\\
\hline
 \end{tabular}
  
 \end{minipage} 
 \begin{minipage}{.5\textwidth}
 \centering
 \scalefont{0.85} 
\caption{Number of news texts} 
\centering
\begin{tabular}{|p{15mm}|p{25mm}|p{10mm}|}
\hline
\textbf{News}    & \textbf{Number of Texts} & \textbf{\%} \\
\hline
Fake    & 3,600         & 50.0  \\
True  	& 3,600         & 50.0 \\
\hline
 \end{tabular}
 \end{minipage}%
\end{table}
 
The Fake.Br corpus was composed in a semi-automatic way. The fake news were collected from sites that gather such content and the true ones were extracted from major news agencies in Brazil, as G1, Folha de São Paulo and Estadão portals. A crawler searched in the corresponding web pages of these agencies for keywords of the fake news, which were nouns and verbs that occurred in the fake news titles and the most frequent words in the texts (ignoring stopwords). The authors have performed a final manual verification to guarantee that the fake and true news were in fact subject-related.

\section{Linguistic Features}
Most of the false statements present linguistic features that are different in relation to the true statements. According to \cite{Conroy:2015:ADD:2857070.2857152}, most liars use their language strategically to avoid being caught. In spite of the attempt to control what they area saying, language ``leakage'' occurs with certain verbal aspects that are hard to monitor, such as frequencies and patterns of pronouns, conjunctions, and negative emotion word usage \cite{feng-hirst-2013-detecting}. 

In this section, we aim at understanding relevant linguistic properties of fake and true statements in Brazilian news. We used Python 3.6.9 and the spaCy  \footnote{\url{https://spacy.io/}} library to automatically annotate \footnote{\url{https://spacy.io/api/annotation}} the corpus. We divided our analysis in two main groups: word-level and sentence-level analyses. In the first group, we analyzed the occurrence patterns of (i) sentiment and emotion words, (ii) part-of-speech tags, (iii) pronoun classification, (iv) named-entity recognition and (v) punctuation behavior. In the second group, we evaluated the number of sentences in fake and true news and the average of words for each sentence. We also analyzed the occurrence of clausal relations in syntactical dependency trees on fake and true statements. We present the results in what follows.

\subsection{Word-Level Analysis}
In the word-level analysis, our goal is to identify differences among word usage behavior and variations in fake and true news. 

\subsubsection{Sentiment and Emotion Words.} According to \cite{DBLP:conf/cogsci/ZipitriaSS17}, deception language involves negative emotions, which are expressed in language in terms of psychological distance from the deception object. The psychological distance and emotional experience reflect an attempt to control the negative mental representation. Therefore, we have identified the incidence of sentiment and emotion words in fake and true statements. We used the sentiment lexicon for Portuguese Sentilex-PT \cite{sentilexpt2012} and WordNetAffect.BR \cite{WordNetAffectBR2008} to account for the sentiment words. Table 4 shows the results. Note that the incidence of sentiment and emotion words in fake news overcame the ones in true news, except for surprise emotion.

In the fake news, we have observed a difference of 11,49 \% and 12,57 \% in positive and negative sentiment when compared to the true news; for joy, sadness, fear, disgust, angry and surprise emotions, the difference amounts to 14,49 \%, 28,92 \%, 7,85 \%, 6,79 \% 12,80 \% and 0,95 \% when compared to the true news. Therefore, we evidence that in our corpus the fake statements presented more negative and positive sentiments and emotions than true statements, confirming what some relevant literature \cite{CluesDeception2003} \cite{CoroyetAll2015} \cite{NewmanetAll2003} \cite{Rubin:2015:DDN:2857070.2857153} have found, i.e., that dishonest texts have more negative than positive sentiments and emotions.

 \begin{table}[!ht]
\scalefont{0.85} 
\caption{Word-level sentiment and emotion occurrence} 
\centering
\begin{tabular}{|p{18mm}|p{18mm}|p{18mm}|}
\hline
\textbf{Sentences} & \textbf{True News} & \textbf{Fake News}  \\ 
\hline
Positive    & 103,376   & 115,260  \\
Negative    & 102,54    & 115,431      \\
Joy         & 4,941     & 5,657  \\
Sadness     & 2,596     & 3,347    \\
Fear        & 1,757     & 1,895     \\
Disgust     & 1,561     & 1,667      \\
Angry       & 2,865     & 3,232     \\
Surprise    & 423       & 419    \\
\hline
Total       & 642,636   & 665,489 \\
\hline
\end{tabular}
\end{table}

\subsubsection{Part-of-Speech.} The growing body of research suggests that we may learn a great deal about people's underlying thoughts, emotions, and reasons by counting and categorizing the words they use to communicate. For \cite{NewmanetAll2003}, several aspects of linguistic style, such as pronoun usage, preposition and conjunctions that signal cognitive work, have been linked to a number of behavioral and emotional outcomes. To exemplify, in \cite{Stirman2001WordUI}, the authors identified that poets who use a high frequency of self-reference but a lower frequency of other-reference in their poetry were more likely to commit suicide than those who showed the opposite pattern. 

In this present study, we extracted the frequency of part-of-speech in our corpus in order to examining the grammatical manifestations of false behavior in text. The obtained results for part-of-speech occurrence is shown in Table 5. The results show an impressive increase on the number of interjections in fake news compared to the true news. We must also point out that, for many authors, it is clear that interjections do not encode concepts as nouns, verbs or adjectives do. Interjections may and do refer to something related to the speaker or to the external world, but their referential process is not the same as that of lexical items belonging to the grammatical categories mentioned, as the referents of interjections are difficult to pin down \cite{Padilla2009}. Similarly, the use of space character has shown a relevant occurrence difference. We found 25,864 spaces in fake news and 3,977 spaces in true news. Furthermore, in true statements, the use of the NOUN category is 9,79 \% larger than in fake statements. The verbal use is also 13,81 \% more frequent in the true statements.

 \begin{table}[!htb]
\scalefont{0.85} 
\caption{Part-of-speech occurrence} 
\centering
\begin{tabular}{|p{5mm}|p{15mm}|p{38mm}|p{18mm}|p{18mm}|}
\hline
\textbf{N.} & \textbf{Label} & \textbf{Definition} & \textbf{True News} & \textbf{Fake News}  \\ \hline
1 & NOUN    & noun                         & 140,107   & 127,609     \\
2 & VERB   & verb                          & 86,256    & 98,168      \\
3 & PROPN  & proper noun                   & 109,501    & 98,757       \\
4 & ADP     & adposition                   & 109,613    & 92,166     \\
5 & ADJ     & adjective                    & 33,433    & 32,535      \\
6 & DET     & determiner                   & 77,660    & 83,169  \\
7 & ADV     & adverb                       & 25,384    & 31,534      \\
8 & \textbf{SPACE}  & \textbf{space}       & \textbf{3,977}    & \textbf{25,864}     \\
9 & PRON   & pronoun                       & 20,994    & 24,348      \\
10 & AUX     & auxiliary                   & 13,529    & 16,999       \\
11 & CCONJ   & coordinating conjunction    & 17,263    & 16,352       \\
12 & NUM     & numeral                     & 16,951    & 12,596     \\
13 & SCONJ  & subordinating conjunction    & 8,870    & 12,392      \\
14 & SYM    & symbol                       & 10,065    & 9,458    \\
15 & OTHER  & other                        & 2,684    & 3,113  \\
16 & \textbf{INTJ}    & \textbf{interjection}                & \textbf{66}    & \textbf{220}   \\
17 & PART   & particle                     & 29    & 23    \\
\hline
\end{tabular}
\end{table}

\subsubsection{Pronouns.} Several studies on deception show that the use of the first-person singular is a subtle proclamation of one's ownership of a statement. In other words, liars tend to distance themselves from their stories and avoid taking responsibility for their behavior \cite{Friedman1990}. Therefore, deceptive communication should be characterized by fewer first-person singular pronouns (e.g., I, me, and my) \cite{NewmanetAll2003}. In addition, when people are self-aware, they are more ``honest'' with themselves \cite{Carver1981} \cite{DuvaletWicklund1972} \cite{Vorauer1999SelfAwarenessAF} and self-reference increases \cite{DavisEtBrock1975}. 

In accordance with deception literature, we investigate the pronoun behavior in our corpus. We identify the occurrence for first, second and third persons of singular and plural pronouns. Table 6 exhibits the results. Surprisingly, the pronoun occurrence in fake news overcame the ones in true news, except in the 3rd person singular (tonic oblique). An unusual behavior, considering the literature on deception, may be noted on the 1st person singular (subject). In fake statements, there has been a jump in the occurrence of the ``eu'' pronoun (1,097) related to true statements (495). 3rd person singular (subject) and 3rd person singular (unstressed oblique) represent 34,16 \% and 42,80 \% respectively on the total occurrence of pronouns in the corpus for the fake news. Differently, for true news, the 3rd person singular (subject) and 3rd person singular (unstressed oblique) represent 36,88\% and 47,74 \% respectively. In other words, the 3rd person occurrence in true news overcame fake news considering the total occurrence of pronouns in the corpus. 

 \begin{table}[!htb]
 \scalefont{0.85} 
\caption{Pronoun occurrence} 
\centering
\begin{tabular}{|p{5mm}|p{58mm}|p{18mm}|p{18mm}|p{18mm}|}
\hline
\textbf{N.} & \textbf{Pronoun Classification} & \textbf{Example} & \textbf{True News} & \textbf{Fake News}  \\ 
\hline
1 & \textbf{1st person singular (subject)}      & \textbf{eu}       & \textbf{495}  & \textbf{1,097} \\
2 & 1st person singular (unstressed oblique)    & me                & 233       & 447   \\
3 & 1st person singular (tonic oblique)         & mim               & 39        & 87    \\
\hline
4 & 2nd person singular (subject)               & você, tu          & 390       & 683  \\
5 & 2nd person singular (unstressed oblique)    & te                & 4          & 24   \\
6 & 2nd person singular (tonic oblique)         & ti, contigo       & 2          &  2   \\
\hline
7 & 3rd person singular (subject)               & ele, ela          & 3,344     & 4,006  \\
8 & 3rd person singular (Unstressed oblique)    & se, o, a, lhe     & 4,329     & 5,019  \\
9 & 3rd person singular (tonic oblique)         & si, consigo       & 44        & 41  \\
\hline
10 & 1st person plural  (subject)               & nós               & 52        & 71      \\
\hline
11 & 2nd person plural (subject)                & vocês             & 7       & 26     \\
\hline
12 & 3rd person plural (subject)                & eles,elas         & 128     & 222  \\
\hline
\end{tabular}
\end{table}

We must also point out that we found that the occurrence differences of 2nd person plural (unstressed oblique) and (tonic oblique) pronouns in fake and true news are statistically irrelevant. 

\subsubsection{Named-Entity Recognition.} According to \cite{Meibauer2018}, most scholars in the field of deception research seem to accept standard truth-conditional. In addition, semantic assumptions on deception are rarely made explicit \cite{Zimmermann2011}. Moreover, the implicit content extraction is a hard task in natural language processing area, as \cite{Vargas&Pardo2018} comments. Nevertheless, we propose a superficial semantic analysis based on named-entity recognition categories. Table 7 shows the results.

\begin{table}[!htb]
\caption{Named-entity occurrence} 
\centering
\scalefont{0.85} 
\begin{tabular}{|p{25mm}|p{15mm}|p{20mm}|p{20mm}|}
\hline
\textbf{Named-Entity} & \textbf{Label} & \textbf{True News} & \textbf{Fake News}  \\ 
\hline
Person              & (PER)         & 19,398  &     22,151   \\
Localization        & (LOC)         & 19,232  &     15,250   \\
Organization        & (ORG)         & 9,503   &     8,851    \\
Miscellaneous       & (MISC)        & 8,427   &     9,119    \\
\hline
\end{tabular}
\end{table}

Based on the obtained results, one may see that the true statements present larger number of localization occurrences (LOC) than fake statements. Otherwise, the fake statements overcome in larger number of person occurrences (PER) when compared to the true statements. Organization (ORG) has occurred more frequently in true statements, while miscellaneous (MISC) in fake statements.
 
\subsubsection{Punctuation.} \cite{CluesDeception2003} assumes that punctuation pattern could distinguish fake and true texts. Consequently, the punctuation behavior would be a ``clue to deception''. We evaluate the occurrence of each punctuation mark. The obtained data is shown in Table 8. In agreement with the literature, our results show a noticeable change among the punctuation setting in fake and true news. Note that in fake statements there has been a expressive use of interrogation, exclamation, end point, double quotes, two points, three consecutive points, square brackets, bar and asterisks. For the true news, we observed the larger use of comma, single trace, single quotes and two consecutive points. We also point out that the ``Error'' label consists on annotation mistakes. 

\begin{table}[!htb]
\caption{Punctuation occurrence} 
\scalefont{0.85} 
\centering
\begin{tabular}{|p{43mm}|p{18mm}|p{18mm}|}
\hline
\textbf{True News} &\textbf{True News} & \textbf{Fake News}  \\ 
\hline
\textbf{Comma}              & \textbf{43,244}    & \textbf{31,610}  \\
\textbf{End point}          & \textbf{26,311}    & \textbf{31,911}  \\
\textbf{Double quotes}      & \textbf{5,004}     &  \textbf{17,200}    \\
Parentheses                 & 12,170             & 11,027  \\
\textbf{Two points}         & \textbf{1,674}     & \textbf{5,138} \\
\textbf{Interrogation}      & \textbf{687}       & \textbf{1,808}    \\ 
\textbf{Exclamation}        & \textbf{277}       & \textbf{2,226}    \\ 
\textbf{Square brackets}    & \textbf{237}       & \textbf{4,638}     \\ 
\textbf{Three consecutive points} & \textbf{71}        & \textbf{2,746}      \\
\textbf{Two consecutive point}    & \textbf{1,619}     & \textbf{127}     \\
Error                             & 1,389              &  1,320 \\
\textbf{Single trace}             & \textbf{910}       & \textbf{13}  \\
Long trace                        & 80                 & 46 \\
\textbf{Asterisk}                 & \textbf{91}        & \textbf{650}   \\
\textbf{Bar}                      & \textbf{123}       & \textbf{358}    \\
\textbf{Single quotes}            & \textbf{443}       &  \textbf{3}     \\
\textbf{Four consecutive points}  & \textbf{0}         & \textbf{54} \\
Circle                            & 4                  & 4 \\
Double trace                      & 12                 & 0 \\
Five consecutive points           & 0                  & 5 \\  
\hline
\end{tabular}
\end{table}

\subsection{Sentence-level Analysis}
According to the standard philosophical definition of lying, the intention to deceive is an important aspect of deception \cite{Augustine1952}  \cite{Kupfer1982} \cite{Williams2002} \cite{Bok1978}. In order to provide an initial understand on dishonest intents in text, we analyze the sentence structure in true and fake statements. In order to achieve that, we evaluate the number of sentences and the average number of words, which is shown in Table 9. 

\begin{table}[!htb]
\scalefont{0.85} 
\caption{Sentence-level analysis} 
\centering
\begin{tabular}{|p{20mm}|p{20mm}|p{20mm}|}
\hline
\textbf{Sentences} & \textbf{True News} & \textbf{Fake News}  \\ 
\hline
Total   & 43,066   & 50,355      \\
Avg of words       & 15,45    & 13,24     \\
\hline
\end{tabular}
\end{table}

Based on data displayed by Table 9, we may note that, in fake statements, there are 14,47\% more sentences than in true statements. It is interesting to realize that, despite the greater number of sentences in fake news, the average number of words by sentence is smaller than in true news. 

According to \cite{Conroy:2015:ADD:2857070.2857152}, analysis of word usage is often not enough for deception prediction. Deeper language structure (syntax) must also be analyzed to predict instances of deception. Therefore, in order to investigate anomalies or divergences in syntactic structure of false and true statements, we also analyzed the dependency relation occurrences in our corpus. We present the results in the Table 10. 

A dependency tree, according to \cite{Jurafsky:2009:ESM:1620754.1620847}, is a syntactic structure corresponding to a given natural language sentence. This structure represents hierarchical relationships between words. Figure 1 shows a dependency tree example. Notice that the relations among the words are illustrated above the sentence with directed labeled arcs from heads to dependents. According to \cite{JurafskyM09}, we call this a typed dependency structure because the labels are drawn from a fixed inventory of grammatical relations. It also includes a \textit{root} node that explicitly marks the root of the tree, i.e., the head of the entire structure. For the interested reader, the Universal Dependencies project \cite{nivre-etal-2016-universal} provides an inventory of dependency relations that are cross-linguistically applicable.  

\begin{figure}[!htb]
\centering                          
\includegraphics[width=1.0\textwidth]{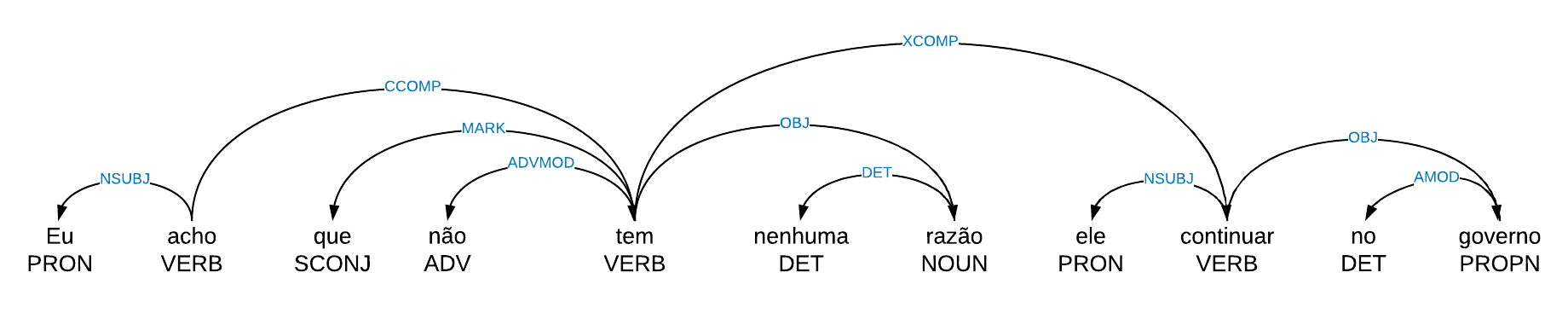}
\caption{Dependency tree example}
\end{figure}

Figure 1 shows the syntactical dependency structure for the following sentence extracted from our corpus: \textit{Eu acho que não tem nenhuma razão ele continuar no governo.} (``I think there is no reason for him to remain in the government''). The NSUBJ relation identifies the subject; CCOMP identifies the complement of the main verb; ADVMOD identifies the adverb modifier; MARK is the word introducing a finite clause subordinate to another clause; OBJ identifies the direct object; DET identifies determinants; AMOD exhibits the adjectival modifier of a noun phrase (NP); and XCOMP consists of an open clausal complement for a verb \footnote{The typed dependency manual in available at \url{https://nlp.stanford.edu/software/dependencies_manual.pdf}}. 

Based on the obtained results (see Table 10), in an initial analysis, we found a relevant difference among the syntactic structures in fake and true news. For example, one may notice a significant difference on the occurrence of CASE, DET, NMOD, ROOT, FLAT:NAME, SUBJ, OBJ, OBL, ADVCL, AUX, PARATAXIS and CSUBJ structures. In the future, we intend to perform a deeper syntactical analysis of the dependency trees, looking for argument structure differences, for instance.

\begin{table}[!htb]
\scalefont{0.80} 
\caption{Clausal dependency relations occurrence} 
\centering
\begin{tabular}{|p{5mm}|p{23mm}|p{55mm}|p{17mm}|p{17mm}|}
\hline
\textbf{N.} & \textbf{Label} & \textbf{Definition} & \textbf{True News} & \textbf{Fake News}  \\ 
\hline
1 & \textbf{CASE}   & \textbf{case marking}             & \textbf{106,964}   & \textbf{89,177} \\
2 & \textbf{DET}     & \textbf{determiner}              & \textbf{71,070}    & \textbf{78,013}  \\
3 & AMOD    & adjectival modifier      & 29,486    & 29,580       \\
4 & \textbf{NMOD}   & \textbf{nominal modifier }        & \textbf{62,406}   & \textbf{50,913}   \\
5 & \textbf{ROOT}    & \textbf{root}                     & \textbf{43,055}    & \textbf{50,035}       \\
6 & \textbf{FLAT:NAME} & \textbf{flat multiword expression (name)}           & \textbf{45,955}    & \textbf{40,025}  \\
7 & \textbf{SUBJ}   & \textbf{nominal subject}           & \textbf{43,091}    & \textbf{49,321}     \\
8 & \textbf{OBJ}    & \textbf{object}                    & \textbf{39,787} & \textbf{45,063}     \\
9 & \textbf{OBL}     & \textbf{oblique nominal}           & \textbf{38,901}    & \textbf{33,550}     \\
10 & ADVMOD & adverbial modifier        & 22,741    & 28,834      \\
11 & CONJ   & conjunct                  & 21,316    & 20,907      \\
12 & APPOS  & appositional modifier     & 21.146    & 20,587       \\
13 & CC     & coordinating conjunction  & 18,603    & 17,586      \\
14 & MARK   & marker                    & 17,108    & 19,932      \\
15 & ACL    & clausal modifier of noun (adjectival clause)           & 14,239    & 13,072    \\
16 & NUMMOD & numeric modifier          & 10,034    & 8,427      \\
17 & COP    & copula                    & 8,767     & 11,359  \\
18 & \textbf{ADVC}L  & \textbf{adverbial clause modifier} & \textbf{8,210}     & \textbf{9,437}     \\
19 & ACL:RELCL & relative clause modifier          & 8,177     & 7,720     \\
20 & CCCOMP & clausal complement        & 7,610     & 9,380      \\
21 & \textbf{AUX}    & \textbf{auxiliary}                 & \textbf{6,902}     & \textbf{9,989}     \\
22 & XCOMP  & open clausal complement   & 6,208     & 7,411      \\
23 & AUX:PASS & auxiliary               & 5,931     & 6,485      \\
24 & NSUBJ:PASS & passive nominal subject        & 5,574     & 6,014      \\
25 & DEP    & unspecified dependency           & 3,017     & 2,357     \\
26 & EXPL   & expletive           & 2,139     & 2,874      \\
27 & NMOD:NPMOD & nominal modifier        & 2,055     & 2,093      \\
28 & OBL:AGENT  & agent modifier       & 1,329     & 1,084      \\
29 & COMPOUND   & compound       & 1,251     & 1,119      \\
30 & \textbf{NMOD:TMOD}  & \textbf{temporal modifier}     & \textbf{1,193}     & \textbf{368 }     \\
31 & FIXED      & fixed multiword expression       & 1,135     & 1,259      \\
32 & \textbf{PARATAXIS}  & \textbf{parataxis}      & \textbf{934}      & \textbf{1,491}      \\
33 & \textbf{CSUBJ}     & \textbf{clausal subject}       & \textbf{769}      & \textbf{1,053}      \\
34 & IOBJ       & indirect object       & 328       & 466      \\
35 & FLAT:FOREIGN & foreign words     & 13        & 15      \\
\hline
\end{tabular}
\end{table}

\section{Final Remarks and Future Work}

We know that language may be used to deceive and confuse people. The current context of social media usage is unique, with diversity in format, and relatively new. However, lying and deceiving have been at play in other forms of human communication for ages \cite{Rubin2017}. In this paper, we presented a study on the statements in true and fake news for Brazilian Portuguese. We performed an empirical linguistic-based analysis over the Fake.Br corpus. We automatically annotated a set of linguistic features in order to investigate actionable inputs and relevant differences among fake and true news. Based on the obtained results, we found that fake and true news present relevant differences in structural, lexical, syntactic and semantic levels.For future work, we intend to deepen our investigation of syntactical behavior and to explore discourse markers and sophisticate machine learning techniques in order to provide deception detection classifiers for different tasks, such as fake news and opinion spam detection in several languages.

\section*{Acknowledgments}
The authors are grateful to CAPES and USP Research Office (PRP 668) for supporting this work.

%
%
\bibliographystyle{splncs04}
\bibliography{bib_file.bib}

\end{document}